\documentclass{article}
\usepackage{spconf,amsmath,graphicx}

\usepackage{booktabs}
\usepackage{adjustbox}
\usepackage{multirow}
\usepackage{pifont}
\usepackage{subfigure}
\usepackage{xspace}
\makeatletter
\DeclareRobustCommand\onedot{\futurelet\@let@token\@onedot}
\def\@onedot{\ifx\@let@token.\else.\null\fi\xspace}

\def\etal{\emph{et al}\onedot}
\usepackage{hyperref}
\usepackage[capitalize]{cleveref}
\title{Beyond Shadows: A Large-Scale Benchmark and Multi-Stage Framework for High-Fidelity Facial Shadow Removal}
\name{\small
Tailong Luo$^{1}$ \qquad 
Jiesong Bai$^{2}$ \qquad 
Jinyang Huang$^{3}$ \qquad 
Junyu Xia$^{2}$ \qquad 
Wangyu Wu$^{4}$ \qquad 
Xuhang Chen$^{5^\dagger}$\thanks{{$^\dagger$} Corresponding Author}}
  
\address{\small $^{1}$ New York Institute of Technology, $^{2}$ Shanghai University, $^{3}$ Central South University\\\small$^{4}$ Xi'an Jiaotong-Liverpool University, $^{5}$ Huizhou University}
\begin{document}
%
\maketitle
\begin{abstract}
Facial shadows often degrade image quality and the performance of vision algorithms. Existing methods struggle to remove shadows while preserving texture, especially under complex lighting conditions, and they lack real-world paired datasets for training. We present the Augmented Shadow Face in the Wild (ASFW) dataset, the first large-scale real-world dataset for facial shadow removal, containing 1,081 paired shadow and shadow-free images created via a professional Photoshop workflow. ASFW offers photorealistic shadow variations and accurate ground truths, bridging the gap between synthetic and real domains. Deep models trained on ASFW demonstrate improved shadow removal in real-world conditions. We also introduce the Face Shadow Eraser (FSE) method to showcase the effectiveness of the dataset. Experiments demonstrate that ASFW enhances the performance of facial shadow removal models, setting new standards for this task.
\end{abstract}

\section{Introduction}
\label{sec:intro}
Facial shadow removal is crucial in computer vision and image enhancement, significantly impacting facial analysis systems. Shadows result from light interactions, facial geometry, and occlusions, which affect image quality and recognition accuracy. Removing shadows while maintaining facial features restores natural appearances. Most approaches rely on supervised learning; self-supervised and unsupervised methods are less explored~\cite{liu2022blind,jin2021dc}. Large-scale paired datasets are crucial but scarce, limiting model performance. Real-world paired data collection is challenging due to technical difficulties, privacy, diversity, and ethical constraints, while synthetic datasets face domain gaps~\cite{zhang2020portrait}. High-quality datasets are urgently needed for effective shadow removal. The \textbf{Augmented Shadow Face in the Wild (ASFW)} dataset, a large benchmark with shadow/shadow-free image pairs, addresses this need. The SFW dataset~\cite{liu2022blind} lacks aligned counterparts for supervised training, so ASFW extends it by adding realistic shadows while ensuring pixel alignment~\cite{liu2022blind}. ASFW includes 1,081 paired images across diverse conditions, serving as a robust benchmark. Additionally, the Face Shadow Eraser (FSE) framework uses shadow-aware mask generation, coarse removal, and facial refinement for efficient shadow removal.
\begin{figure}[ht]
\centering
\includegraphics[width=\linewidth]{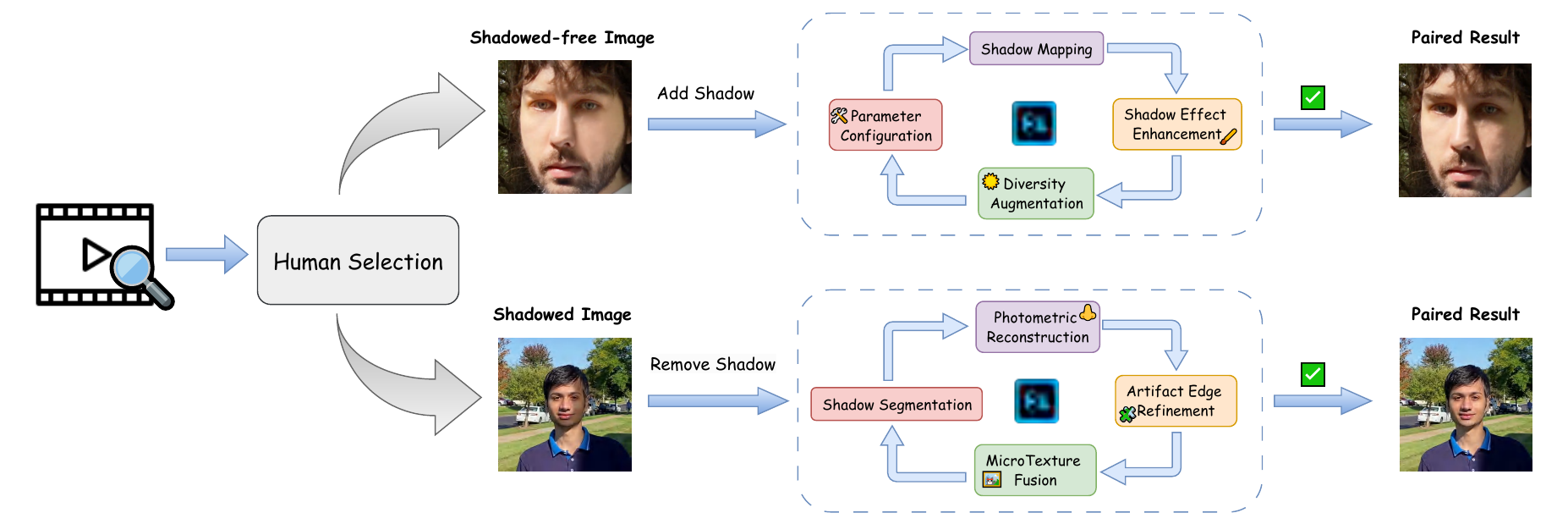}
\caption{Construction process of the ASFW dataset. The bidirectional workflow includes shadow synthesis (top) and shadow removal (bottom), leveraging Adobe Photoshop tools for photorealistic alignment.}
\label{fig1}
\end{figure}
The main contributions of this study are three-fold. 1) \textbf{ASFW Dataset}: The first manually annotated benchmark offering 1,081 aligned shadow/shadow-free pairs with photorealistic variations, addressing critical data scarcity issues in the facial shadow removal problem. 2) \textbf{FSE Framework}: A simple but effective three-stage architecture combining mask-guided detail preservation, multi-scale context aggregation, and symmetry-aware refinement, achieving good performance on both synthetic and real-world benchmarks. 3) \textbf{Empirical Validation}: Extensive experiments demonstrate that our ASFW dataset can help solve the facial shadow removal problem better than existing methods and datasets.
\section{ASFW dataset}
\label{sec:data}

In this section, we first show the challenges in facial shadow dataset construction. Then, we provide the details about the construction of our dataset and the differences between our dataset and the existing dataset. 

\subsection{Challenges in Dataset Construction}

Creating a high-quality paired dataset for facial shadow removal involves technical and perceptual challenges. Realistic shadow synthesis needs fine-tuned optical parameters like flow direction, edge softness, and opacity to mimic various lighting and occlusion effects. Shadows must align with 3D facial features, especially around critical areas like the nasal bridge and zygomatic arches, requiring skill in image editing and knowledge of facial anatomy. Effective shadow removal must eliminate light-induced artifacts while maintaining facial textures. Unlike existing datasets prone to over-smoothing, our multi-stage manual refinement ensures natural appearance and structural consistency, albeit at higher costs, with steps like detailed boundary annotation and illumination normalization overseen by experts.

\subsection{Data Construction Pipeline}

To address these challenges, we leverage the SFW video dataset~\cite{liu2022blind}, which mitigates privacy concerns inherent in facial data collection. We extract representative frames exhibiting diverse facial poses, expressions, and lighting conditions to ensure dataset comprehensiveness. Our curation process follows a bidirectional approach—adding realistic shadows to shadow-free images and meticulously removing shadows from shadowed images—using Adobe Photoshop's professional editing capabilities. 

\subsubsection{Shadowed Image Creation Pipeline}

The shadow generation process has four stages for photorealism: \textbf{Parameter Configuration:} Calibrate optical properties by adjusting brush flow (10-30\%) and opacity (15-45\%) in Adobe Photoshop, fine-tuning shadows to mimic natural transitions, especially in penumbra areas. \textbf{Shadow Mapping:} Align shadows with facial landmarks using templates, focusing on areas prone to cast shadows like the nasal bridge, orbital areas, and zygomatic arches. \textbf{Shadow Effect Enhancement:} Use dual-edge rendering by adjusting brush diameter (5-15 pixels for hard shadows and 25-50 pixels for soft shadows) and pressure-sensitive opacity to ensure smooth transitions and refined boundaries. \textbf{Diversity Augmentation:} Simulate various occlusion scenarios including shadows from hair, hats, and hands. Incorporate micro-shadows from facial textures such as wrinkles and pores for added realism.

\subsubsection{Shadow-Free Image Creation Pipeline}

Shadow removal involves a four-stage refinement process: \textbf{Shadow Segmentation:} The Lasso Tool with adaptive feathering precisely selects shadow areas, adjusting feather radius for smooth transitions. \textbf{Photometric Reconstruction:} Brightness and color corrections restore illumination in shadows, with feathering adjustments preventing halos and ensuring tonal consistency. \textbf{Artifact Edge Refinement:} The Spot Healing Brush Tool addresses minor boundary inconsistencies using content-aware sampling for better continuity. \textbf{Micro-Texture Preservation:} Content-Aware Fill and the Clone Stamp Tool fix local imperfections, while the Mixer Brush Tool ensures natural edge blending and retains facial textures.

\begin{figure*}[ht]
\centering
\includegraphics[width=\linewidth]{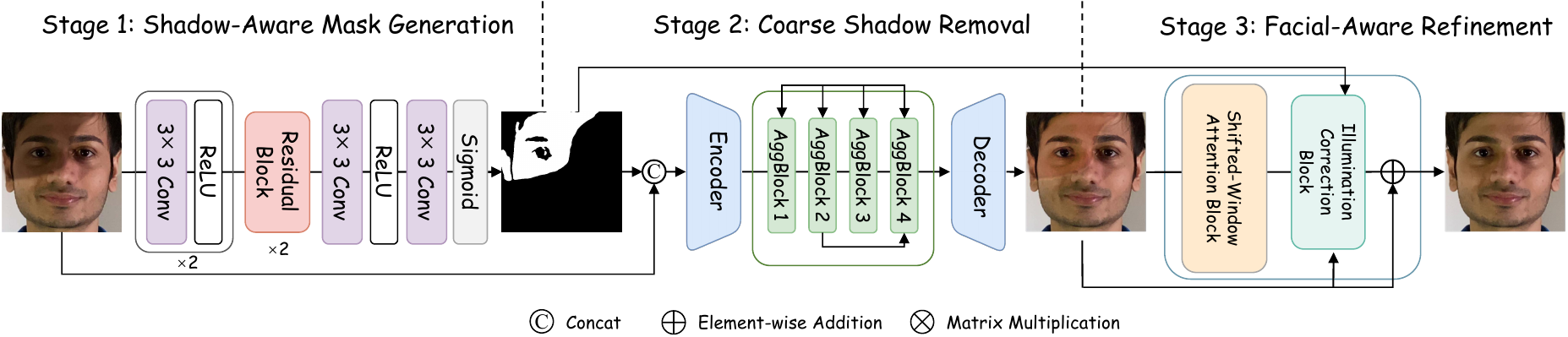}
\caption{Model overview.}
\label{fig2}
\end{figure*}
\section{Method}
We present \textbf{Face Shadow Eraser (FSE)}, a lightweight yet expressive framework for high-fidelity facial shadow removal. FSE adopts a progressive pipeline—shadow-aware localization, coarse occlusion removal, and fine-scale facial refinement—jointly optimizing structural consistency and photometric fidelity. Model complexity analysis is provided in the Supplementary Materials.
\subsection{Overall Framework}
\label{overall}
As illustrated in \cref{fig2}, the FSE architecture progressively removes facial shadows through cascaded stages, formulated as:
\begin{equation}
R = \mathcal{F}_{\rm{refine}} \circ \mathcal{F}_{\rm{coarse}} \circ \mathcal{F}_{\rm{mask}}(I \oplus M),
\end{equation}
where $\oplus$ denotes channel-wise concatenation, and $\circ$ represents function composition. Given an input image $I$ and an optional initial mask $M$. $M$ is generated by thresholding the difference between the target and input images.

\subsection{MaskGuideNet: Shadow-Aware Mask Generation}
\label{subsec:mask}
In our implementation, the shadow mask generation module, called MaskGuideNet, takes a 4-channel input $(I \oplus M)$ where $I$ is the RGB image and $M$ is an optional initial mask. This step yields the refined shadow probability map $M' \in [0,1]^{H \times W}$, providing a robust soft shadow map for subsequent stages.
\begin{equation}
    M' = \operatorname{Sigmoid} \Big(\mathcal{E} \big( \mathcal{D} \big( \mathcal{E}(I\oplus M) \big) \big) \Big),
    \label{eq:mask}
\end{equation}
where $\mathcal{E}/\mathcal{D}$ denotes the feature extraction/residual blocks in $\mathcal{F}_{\rm{mask}}$ and $\operatorname{Sigmoid}$ is the standard sigmoid activation function.

\subsection{CoarseGenNet: Coarse Shadow Removal}
\label{subsec:coarse}
The second stage, CoarseGenNet, casts the coarse shadow removal as:
\begin{equation}
    C = \mathcal{F}_{\rm{coarse}}(I,\,M'),
\end{equation}
where $I$ and $M'$ are concatenated at the input. The network begins with an initial feature extraction, followed by a sequence of four $\operatorname{AggBlock}$ modules that progressively expand the receptive field and fuse learned features. Each $\operatorname{AggBlock}$ internally leverages dynamic convolutional layers with different dilation rates, allowing multi-scale feature extraction without explicit downsampling. 

\subsection{RefineFaceNet: Facial-Aware Refinement} 
\label{subsec:refine}
In the final stage, RefineFaceNet refines the coarse output by integrating global attention and local illumination correction, aiming to eliminate residual shadows while preserving facial detail. The overall formulation is:
\begin{equation}
    R = C + \underbrace{\operatorname{AHSWA}(C)}_{\operatorname{global~attention}} \otimes \underbrace{\operatorname{IRC}(C \oplus M)}_{\operatorname{local~correction}},
    \label{eq:refine}
    \end{equation}
where Adaptive Hierarchical Shift-Window Attention (AHSWA) builds upon the Swin Transformer~\cite{liu2021swin}, applying multi-scale windowed self-attention with alternating regular and shifted window partitioning. Attention within each window is computed using scaled dot-product attention with relative position bias, followed by depthwise convolutions to strengthen local continuity crucial for face structure preservation.

To correct uneven illumination, our Illumination Refinement Component (IRC) employs mask-conditional feature modulation through three consecutive modulation blocks:
\begin{equation}
    \mathbf{C}' = (\gamma(\mathbf{M}') \otimes \mathbf{W}_\gamma + \beta(\mathbf{M}')) \otimes \mathbf{C},
\end{equation}
where $\gamma(\cdot), \beta(\cdot)$ are pointwise convolutional networks conditioned on shadow mask $\mathbf{M}'$, $\mathbf{W}_\gamma$ denotes channel-wise weights, and $\otimes$ represents element-wise multiplication. 

Putting everything together, the RefineFaceNet output is computed as:
\begin{equation}
    R = \underbrace{\operatorname{RefineFaceNet}(C,\, M')}_{\operatorname{AHSWA + IRC}},
\end{equation}
where $R$ is the reconstructed shadow-free image.

\section{Experiment}
\begin{table}[ht] \centering \footnotesize 
\setlength{\tabcolsep}{9pt} 
\renewcommand{\arraystretch}{1.2} 
\caption{Quantitative comparison on ASFW and UCB datasets. The best results are marked in bold.} \adjustbox{width=\linewidth}{ \begin{tabular}{lcccccccc} \toprule \multirow{2}{*}{Method} & \multicolumn{4}{c}{ASFW} & \multicolumn{4}{c}{UCB} \\ \cmidrule(lr){2-5} \cmidrule(lr){6-9} & PSNR $\uparrow$ & SSIM $\uparrow$ & MSE $\downarrow$ & LPIPS $\downarrow$ & PSNR $\uparrow$ & SSIM $\uparrow$ & MSE $\downarrow$ & LPIPS $\downarrow$ \\ \midrule Yang \etal~\cite{yang2012shadow} & 13.77 & 0.711 & 0.046 & 0.230 & 12.83 & 0.538 & 0.054 & 0.234 \\ Guo \etal~\cite{guo2012paired} & 16.00 & 0.771 & 0.037 & 0.196 & 17.80 & 0.712 & 0.022 & 0.150 \\ Gryka \etal~\cite{gryka2015learning} & 10.12 & 0.404 & 0.106 & 0.558 & 10.92 & 0.301 & 0.090 & 0.445 \\ \midrule DeShadowNet~\cite{qu2017deshadownet} & 17.22 & 0.837 & 0.023 & 0.181 & 17.42 & 0.709 & 0.020 & 0.196 \\ ST-CGAN~\cite{wang2018stacked} & 22.57 & 0.854 & 0.008 & 0.126 & 19.11 & 0.746 & 0.013 &0.130 \\ Mask-ShadowGAN~\cite{hu2019mask} & 22.79 & 0.876 & 0.008 & 0.132 & 18.35 & 0.749 & 0.016 & 0.125\\ DHAN~\cite{cun2020towards} & 21.73 & 0.913 & 0.015 & 0.076 & 18.31 & 0.762 & 0.019 & 0.117 \\ LG-ShadowNet~\cite{liu2021shadow} & 21.57 & 0.864 & 0.009 & 0.146 & 18.53 & 0.739 & 0.015 & 0.127 \\ G2R-ShadowNet~\cite{liu2021shadow} & 22.78 & 0.880 & 0.008 & 0.130 & 18.49 & 0.751 & 0.015 & 0.127 \\ DC-ShadowNet~\cite{jin2021dc} & 21.33 & 0.912 & 0.014 & 0.080 & 18.60 & 0.762 & 0.017 & 0.117 \\ SG-ShadowNet~\cite{wan2022style} & 19.47 & 0.754 & 0.017 & 0.252 & 16.27 & 0.711 & 0.026 & 0.148 \\ BMNet~\cite{zhu2022bijective} & 23.65 & 0.927 & 0.009 & 0.069 & 19.78 & 0.774 & 0.012 & 0.104 \\ TBRNet~\cite{liu2023shadow} & 18.19 & 0.848 & 0.017 & 0.150 & 16.51 & 0.683 & 0.023 & 0.161 \\ ShadowFormer~\cite{guo2023shadowformer} & 21.99 & 0.914 & 0.012 & 0.082 & 19.76 & \textbf{0.781} & 0.012 & 0.103 \\ DMTN~\cite{liu2023decoupled} & 19.32 & 0.852 & 0.016 & 0.128 & 18.26 & 0.723 & 0.017 & 0.131\\ ShadowRefiner~\cite{dong2024shadowrefiner} & 21.57 & 0.877 & 0.011 & 0.136 & 19.55 & 0.768 & 0.012 & 0.121 \\ S3RNet~\cite{kubiak2024s3r} & 21.22 & 0.908 & 0.014 & 0.082 & 18.23 & 0.752 & 0.019 & 0.121 \\ RASM~\cite{liu2024regional} & 13.64 & 0.633 & 0.051 & 0.359 & 13.58 & 0.601 & 0.049 & 0.267 \\ HomoFormer~\cite{xiao2024homoformer} & 17.29 & 0.873 & 0.024 & 0.115 & 19.00 & 0.758 & 0.015 & 0.112 \\ \midrule PSM~\cite{zhang2020portrait} & 19.46 & 0.865 & 0.029 & 0.149 & 18.65 & 0.754 & 0.016 & 0.125 \\ BlindShadow~\cite{liu2022blind} & 21.86 & 0.909 & 0.012 & 0.079 & 18.70 & 0.754 & 0.016 & 0.116 \\ Lyu \etal~\cite{lyu2022portrait} & 23.69 & 0.903 & 0.007 & 0.089 & 19.41 & 0.769 & 0.012 & 0.118 \\ FSRNet~\cite{zhang2024frequency} & 21.90 & 0.914 & 0.015 & 0.075 & 18.35 & 0.763 & 0.019 & 0.118 \\ CIRNet~\cite{yu2024portrait} & 21.76 & 0.914 & 0.015 & 0.075 & 18.24 & 0.763 & 0.019 & 0.118 \\ \textbf{Ours} & \textbf{25.45} & \textbf{0.930} & \textbf{0.006} & \textbf{0.066} & \textbf{20.15} & 0.774 & \textbf{0.011} & \textbf{0.103} \\ \bottomrule \end{tabular}} \label{table2} \end{table}

\label{sec:exp}
\subsection{Implementation Details}
We utilize the full FFHQ dataset~\cite{karras2019style}, containing 70,000 high-quality facial images, for model training. Synthetic shadows are generated using methodologies from~\cite{hu2019mask} to simulate realistic shadows. For evaluation, we adopt two benchmark datasets: (1) UCB~\cite{zhang2020portrait}, comprising 100 paired shadowed/shadow-free facial images, and (2) ASFW, our proposed dataset with 1,081 paired samples. These datasets enable comprehensive performance assessment across diverse lighting scenarios. The training loss includes MSE for pixel-level reconstruction, SSIM for structural preservation, and LPIPS for perceptual quality:
\begin{equation}
    \mathcal{L} = \mathcal{L}_{\mathrm{MSE}} + \lambda_1 \cdot \mathcal{L}_{\mathrm{SSIM}} + \lambda_2 \cdot \mathcal{L}_{\mathrm{LPIPS}},
\end{equation}
where we empirically set $\lambda_1$ and $\lambda_2$ to be 0.2.

Experiments are implemented in PyTorch and executed on an Ubuntu 22.04 LTS system with 2 Intel Xeon Gold 5215 CPUs, 256GB RAM, and 2 NVIDIA A800 80G GPUs. All models are trained with a batch size of 8, an initial learning rate of $2 \times 10^{-4}$, and an input resolution of $256 \times 256$ pixels. Inference uses $512 \times 512$ (ASFW) and $256 \times 256$ (UCB) resolutions. Optimization employs the AdamW optimizer ($\beta_1=0.9$, $\beta_2=0.999$) with a cosine annealing scheduler. Data augmentation includes random cropping, horizontal flipping, and rotation.

\begin{table}[ht] 
\centering %
\caption{Ablation study for our method's different modules.} \adjustbox{width=\linewidth}{ \begin{tabular}{ccccccccccc} \toprule \multicolumn{3}{c}{Module} & \multicolumn{4}{c}{ASFW} & \multicolumn{4}{c}{UCB} \\ \cmidrule(lr){1-3} \cmidrule(lr){4-7} \cmidrule(lr){8-11} MaskGuideNet & CoarseGenNet & RefineFaceNet & PSNR $\uparrow$ & SSIM $\uparrow$ & MSE $\downarrow$ & LPIPS $\downarrow$ & PSNR $\uparrow$ & SSIM $\uparrow$ & MSE $\downarrow$ & LPIPS $\downarrow$ \\ \midrule  \ding{55} & \ding{51} & \ding{51} & 22.47 & 0.905 & 0.009 & 0.086 & 19.20 & 0.755 & 0.013 & 0.113 \\ \ding{51} & \ding{55} & \ding{51} & 22.04 & 0.893 & 0.009 & 0.104 & 20.07 & 0.760 & 0.011 & 0.106 \\ \ding{51} & \ding{51} & \ding{55} & 22.93 & 0.886 & 0.008 & 0.112 & 19.99 & 0.753 & 0.011 & 0.111 \\ \ding{55} & \ding{51} & \ding{55} & 22.79 & 0.831 & 0.008 & 0.166 & 19.96 & 0.728 & 0.011 & 0.139 \\ \ding{55} & \ding{55} & \ding{51} & 21.48 & 0.864 & 0.010 & 0.126 & 19.15 & 0.741 & 0.013 & 0.121 \\ \ding{51} & \ding{51} & \ding{51} & \textbf{25.45} & \textbf{0.930} & \textbf{0.006} & \textbf{0.066} & \textbf{20.15} & \textbf{0.774} & \textbf{0.011} & \textbf{0.103} \\ \bottomrule \end{tabular}} \label{table3} 
\end{table}
\subsection{Comparisons with other Methods}
\begin{figure}[ht]
    \centering
    \begin{minipage}[b]{\linewidth}
    \subfigure[Input]{
        \begin{minipage}[b]{0.155\linewidth}
            \includegraphics[width=\linewidth]{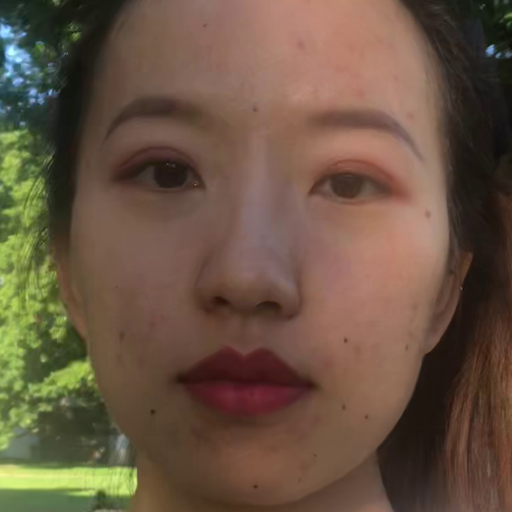}\\ 
            \includegraphics[width=\linewidth]{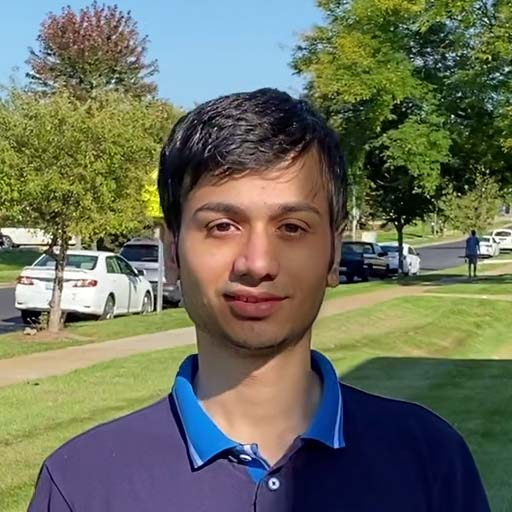}\\
            \includegraphics[width=\linewidth]{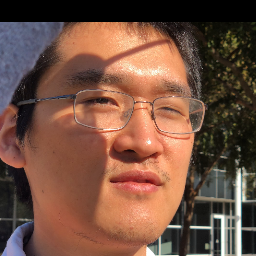}\\
        \end{minipage}
    } \hspace{-5pt} 
    \subfigure[Lyu \etal]{
        \begin{minipage}[b]{0.155\linewidth}
            \includegraphics[width=\linewidth]{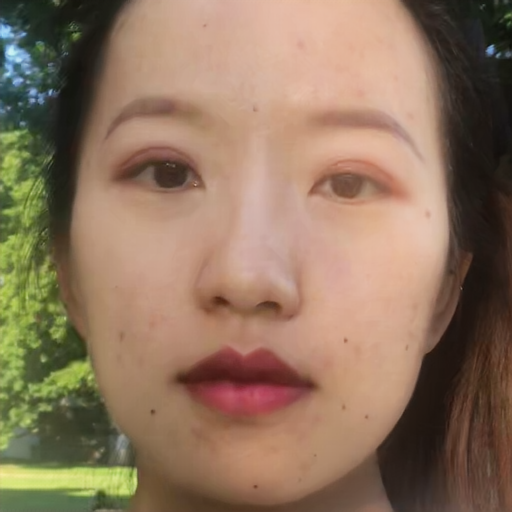}\\ 
            \includegraphics[width=\linewidth]{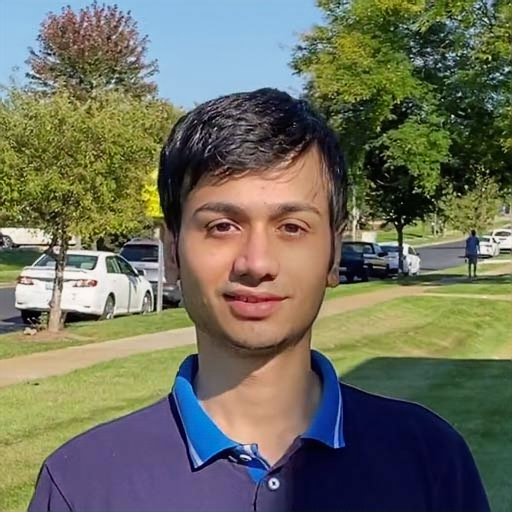}\\
            \includegraphics[width=\linewidth]{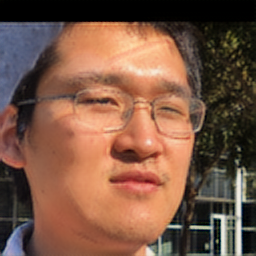}\\
        \end{minipage}
    }\hspace{-5pt}
    \subfigure[FSRNet]{
        \begin{minipage}[b]{0.155\linewidth}
            \includegraphics[width=\linewidth]{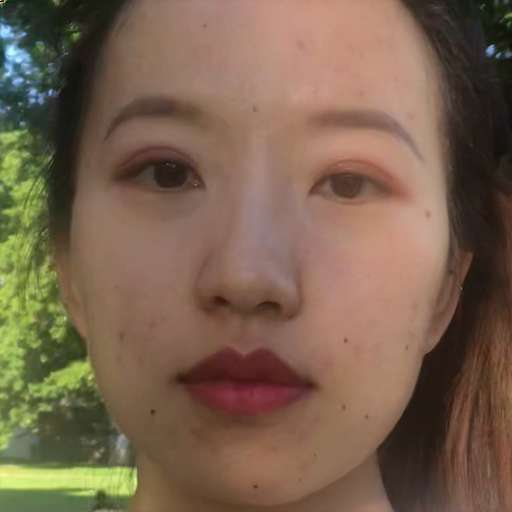}\\ 
            \includegraphics[width=\linewidth]{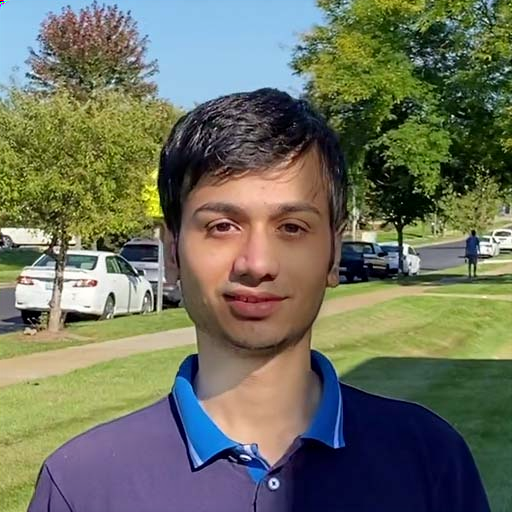}\\
            \includegraphics[width=\linewidth]{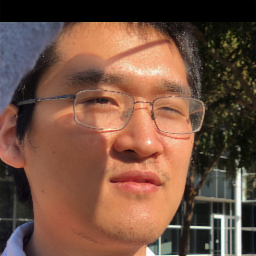}\\
        \end{minipage}
    }\hspace{-5pt}
    \subfigure[CIRNet]{
        \begin{minipage}[b]{0.155\linewidth}
            \includegraphics[width=\linewidth]{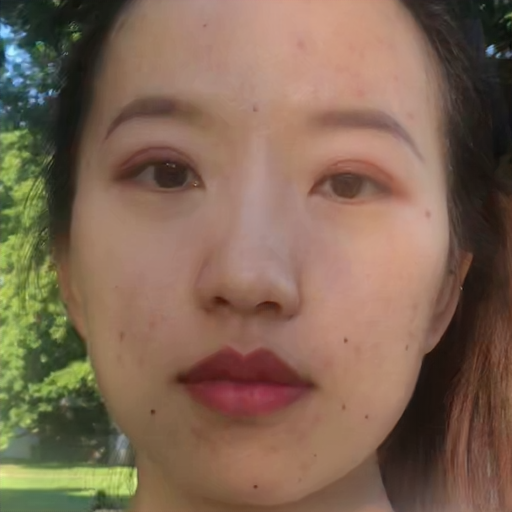}\\
            \includegraphics[width=\linewidth]{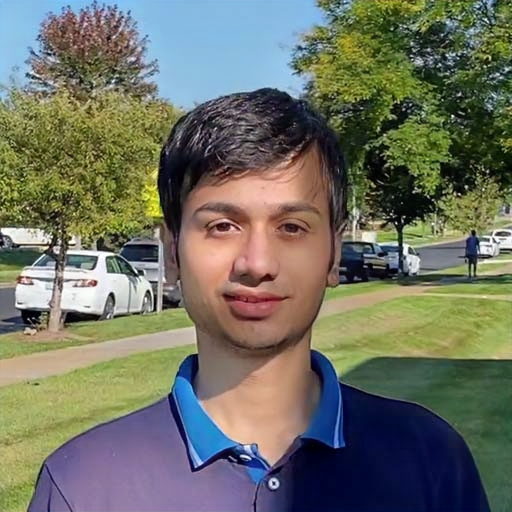}\\
            \includegraphics[width=\linewidth]{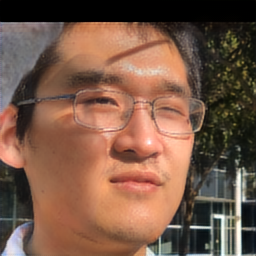}\\
        \end{minipage}
    }\hspace{-5pt}
    \subfigure[Ours]{
        \begin{minipage}[b]{0.155\linewidth}
            \includegraphics[width=\linewidth]{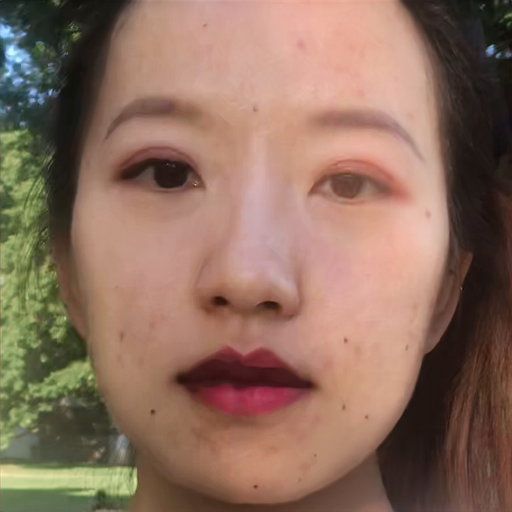}\\ 
            \includegraphics[width=\linewidth]{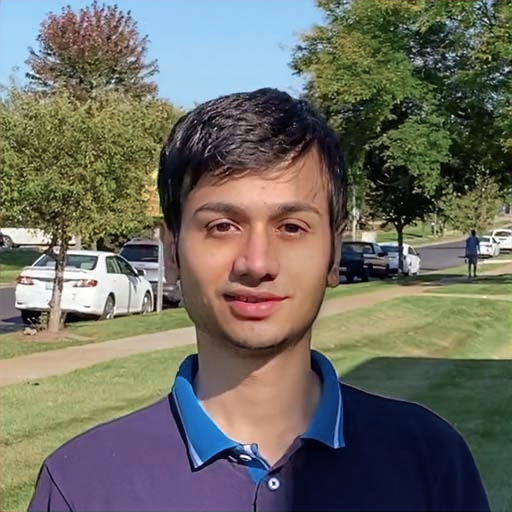}\\
            \includegraphics[width=\linewidth]{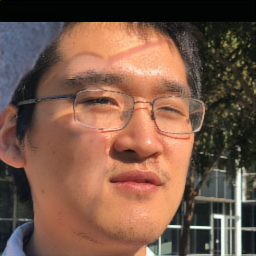}\\
        \end{minipage}
    }\hspace{-5pt}
    \subfigure[GT]{
        \begin{minipage}[b]{0.155\linewidth}
            \includegraphics[width=\linewidth]{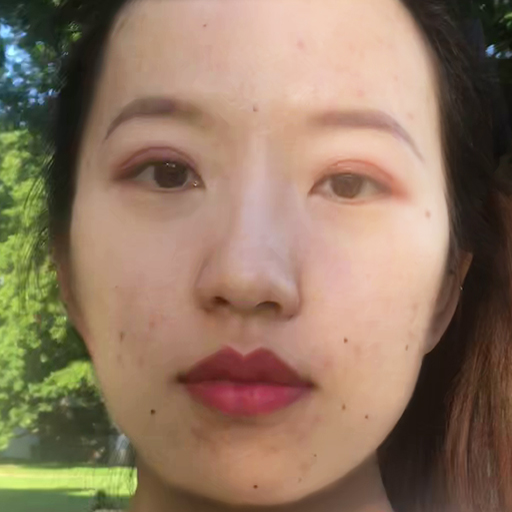}\\ 
            \includegraphics[width=\linewidth]{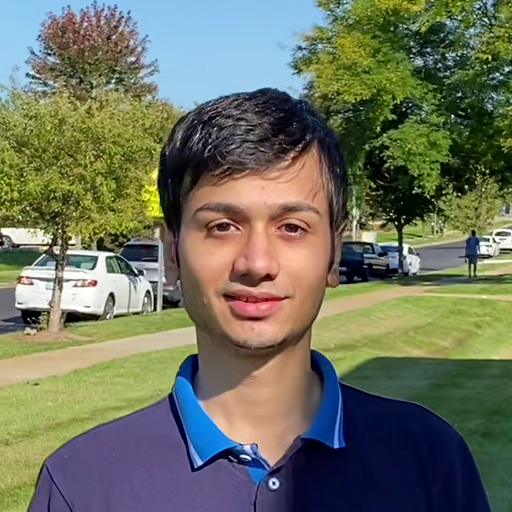}\\
            \includegraphics[width=\linewidth]{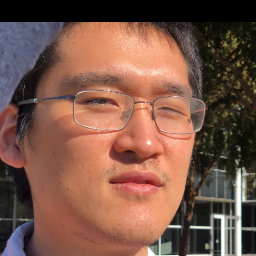}\\
        \end{minipage}
    }
    \end{minipage}
    \caption{Visual comparison of facial shadow removal on ASFW and UCB dataset.} 
    \label{fig:result_comparison}
\end{figure}

\subsubsection{Qualitative Results}
\Cref{fig:result_comparison} presents a qualitative comparison of several face shadow removal methods, including Lyu~\etal~\cite{lyu2022portrait}, FSRNet~\cite{zhang2024frequency}, CIRNet~\cite{yu2024portrait}, and our proposed approach, evaluated in the challenging ASFW dataset and UCB dataset. As shown, while Lyu~\etal and FSRNet achieve reasonable shadow suppression, they still retain subtle shadow boundaries in specific facial regions. CIRNet generates more visually natural results but tends to produce noticeable blurring. Our method demonstrates robust shadow removal with clear preservation of facial structure. Benefiting from the diverse and complex scenarios provided by ASFW—including various shadow types, lighting conditions, and occlusion patterns—our model achieves the closest visual match to ground-truth images. These results validate the utility of ASFW as a high-quality benchmark for face shadow removal. Additional visualizations are provided in the supplementary material.

\subsubsection{Quantitative Results}
Our evaluation uses four metrics: \textbf{PSNR}, \textbf{SSIM}, \textbf{MSE}, \textbf{LPIPS}. Higher PSNR/SSIM and lower MSE/LPIPS indicate better performance. Results in \cref{table2} show the ASFW dataset reveals performance differences among shadow removal methods more effectively than the UCB dataset. ASFW is more challenging and diverse with varied shadow types, lighting, and shadow areas, leading to greater metric variance and a more accurate evaluation of quality and fidelity. Leading methods like SG-ShadowNet~\cite{wan2022style} and ShadowMaskFormer~\cite{guo2023shadowformer} show significant improvements, highlighting ASFW as a strong benchmark for future face shadow removal advancements.
\subsection{Ablation Study}

\cref{table3} presents the quantitative results of our ablation study, evaluating the contributions of MaskGuideNet, CoarseGenNet, and RefineFaceNet to the overall performance of our face shadow removal system. From the results, we observe that the full model, which integrates all three modules, achieves the best performance across all metrics on both datasets. Specifically, it attains the best metrics performance. This demonstrates that each module contributes to improving both the perceptual quality and structural similarity of the restored images. Individually, MaskGuideNet and CoarseGenNet play crucial roles in enhancing PSNR and reducing MSE, while RefineFaceNet significantly refines the final output, as evidenced by the substantial reduction in LPIPS. Notably, removing RefineFaceNet (fourth row) results in a sharp increase in LPIPS, indicating a decline in perceptual quality. Furthermore, the absence of CoarseGenNet (fifth row) leads to the lowest PSNR and SSIM scores, suggesting that a coarse-to-fine structure is essential for effective shadow removal.
\section{Conclusion}
We introduced the ASFW dataset to enhance facial shadow removal, offering 1,081 high-quality shadow/shadow-free image pairs with broad attribute coverage like age, gender, skin tone, pose, and lighting, ensuring diversity and relevance. This dataset aids in training networks to improve image quality and facial analysis stability. Experiments confirm its effectiveness in diverse shadow removal. Additionally, we developed the Face Shadow Eraser (FSE), a three-stage framework that efficiently removes shadows while maintaining image integrity. Our work aims to support future research and development of advanced shadow removal methods.

\section*{Acknowledgment}
This work was supported by the Guangdong Basic and Applied Basic Research Foundation (Grant No. 2024A1515140010).

\bibliographystyle{IEEEbib}
\bibliography{refs}

\end{document}